\documentclass[letterpaper,10pt,conference]{ieeeconf}

\IEEEoverridecommandlockouts
\usepackage{amsmath,amssymb}
\usepackage{booktabs}
\usepackage{cite}
\usepackage{graphicx}

\title{Destination Support Restoration for Finite-Set Multimodal Trajectory Prediction}

\author{Fengrui Liu$^{1}$, Jiajun Peng$^{2}$, Duo Peng$^{3}$, and Feng Liu$^{4,*}$%
\thanks{$^{1}$Fengrui Liu is with the School of Computer Science and Technology, East China Normal University, Shanghai, China.}%
\thanks{$^{2}$Jiajun Peng is with the School of Data Science, University of Science and Technology of China, Hefei, China.}%
\thanks{$^{3}$Duo Peng is with the School of Computer Science and Technology, Tongji University, Shanghai, China.}%
\thanks{$^{4}$Feng Liu is with the School of Psychology, Shanghai Jiao Tong University, Shanghai, China.}%
\thanks{$^{*}$Corresponding author: Feng Liu.}%
}

\begin{document}

\maketitle
\thispagestyle{empty}
\pagestyle{empty}

\begin{abstract}
Robots operating around pedestrians often reason over a finite set of predicted human futures. Repeated online updates can concentrate this limited prediction budget on dominant destinations and leave plausible alternatives underrepresented or absent, removing those alternatives from the finite representation available to downstream decision making. We introduce Destination Support Restoration (DSR), a causal post-selection operator that repairs destination support without retraining the host predictor or increasing the maintained set size. At a repair step, DSR evaluates a temporary destination-stratified candidate bank from the observed prefix, converts candidate evidence into integer target counts, protects representatives of active modes, and reallocates redundant surplus hypotheses to deficient modes. The maintained and returned sets retain exactly $N$ hypotheses, and DSR replaces at most $\lceil\rho N\rceil$ entries. Protected representatives preserve current categorical support; lineage-aware particle filters also preserve surviving resampling ancestors. Each replacement reduces the allocation mismatch to the evidence-driven target by one. On the complete 3,719-trajectory Edinburgh protocol over three seeds, DSR reduces MIF weighted ADE and FDE by 13.36\% and 13.30\% at $N=64$. Paired integrations with CLiFF, PPT, causal GDTS, Social Informer, and PECNet improve both metrics in every evaluated pair. These results show that finite-set support allocation is a useful prediction-side control point when a fixed hypothesis set serves as the interface to downstream systems.
\end{abstract}

\begin{figure}[!t]
    \centering
    \includegraphics[width=\columnwidth]{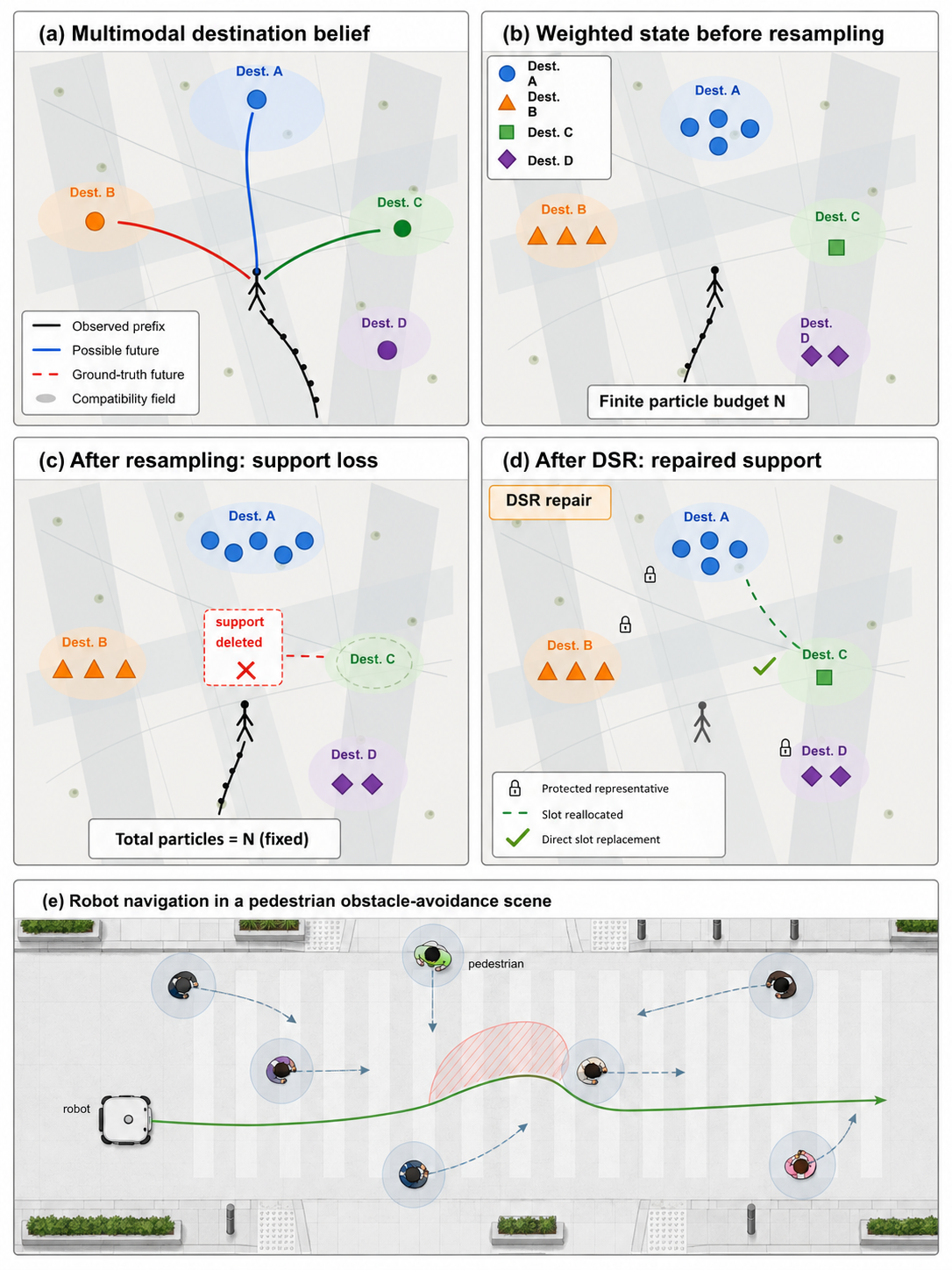}
    \caption{
    Overview of Destination Support Restoration (DSR).
    (a) An observed prefix remains compatible with multiple destination modes.
    (b) A fixed-size maintained set concentrates on dominant modes.
    (c) Resampling may eliminate a still-plausible destination from the finite representation.
    (d) DSR protects surviving support and reallocates surplus slots toward evidence-supported deficits while returning the same number of hypotheses.
    (e)
    The figure illustrates the lineage-aware case; set-based hosts use protected mode representatives instead.
    }
    \label{fig:overview}
    \vspace{-2mm}
\end{figure}

\section{Introduction}
Mobile robots operating around pedestrians must anticipate more than one plausible human future. In practice, a prediction module often exposes a finite hypothesis set to downstream decision making, so the number of represented futures is itself a limited computational resource. If a plausible destination disappears from that set, a downstream planner cannot reason over that alternative even when it remains compatible with the observations. This makes the allocation of a fixed prediction budget relevant in its own right, in addition to the quality of the underlying trajectory generator.

Modern pedestrian predictors represent multimodality with recurrent interaction models, graph architectures, endpoint-conditioned generation, latent variables, and diffusion models~\cite{alahi2016sociallstm,gupta2018socialgan,salzmann2020trajectron,shi2021sgcn,mangalam2020pecnet,mangalam2021ynet,gu2022mid,mao2023led}. Recent methods further improve destination learning, goal-guided diffusion, and probabilistic Transformer prediction~\cite{lin2024ppt,sun2025gdts,jiang2026socialinformer}. These advances improve how candidate futures are generated, but they do not by themselves decide how a finite maintained set should allocate its slots across plausible modes.

Fig.~\ref{fig:overview} shows the failure that motivates this work. A fixed-size set can concentrate on a dominant destination while assigning too few hypotheses, or none, to another plausible destination. The problem is especially visible in recursive predictors: repeated propagation, weighting, and resampling can replicate high-weight particles and remove low-mass categorical destinations from the finite representation~\cite{gordon1993bootstrap,doucet2001sequential}. Once a destination label disappears, continuous perturbations cannot recreate that categorical support. Goal-directed pedestrian filters and MIF-WLSTM therefore expose a gap between the multimodal belief and the support actually represented by the hypotheses available to the trajectory decoder~\cite{rehder2015goal,particke2018improvements,huang2020mif}.

Complete mode extinction is only the limiting case. A plausible destination may still retain one hypothesis while receiving much less of the fixed budget than the current evidence warrants. We refer to both cases as a \textbf{support-allocation mismatch}: the finite set no longer reflects how its slots should be distributed across plausible alternatives. The goal is not to enlarge the output set, but to use the same $N$ slots more effectively by restoring missing support and reallocating redundant support before extinction occurs.

Existing particle-filter remedies address related forms of sample impoverishment. Resample--move methods rejuvenate particles through Markov transitions, KLD sampling changes the particle count, and alternative resampling rules redistribute particle mass~\cite{gilks2001moving,fox2001kld,morelande2011mode,li2012deterministic,corenflos2021dpf}. Random mutation or restart may reintroduce a missing mode, but it does not assign an evidence-driven slot target to each semantic alternative. Diverse trajectory samplers improve one-shot coverage of a frozen predictor~\cite{ma2021lds,bae2022npsn}; our concern is different: how an already finite maintained set should allocate its existing slots online.

We propose \textbf{Destination Support Restoration (DSR)}, a causal operator for this allocation problem. At a repair step, DSR builds a temporary mode-stratified candidate bank from the observed prefix, converts candidate evidence into integer target counts for the $N$ maintained slots, protects representatives of active support, and releases only unprotected hypotheses from target-surplus modes. Donors and recipient candidates are ranked by compatibility energy, and at most $\lceil\rho N\rceil$ slots are reassigned. The host predictor, maintained set size, and returned set size remain unchanged.

DSR separates support repair from the host generator. In a lineage-aware filter such as MIF-WLSTM~\cite{huang2020mif}, it protects the first offspring of every distinct resampling ancestor. Predictors without ancestry protect one representative of every active mode. The same allocation rule acts directly on MIF particles, through GoalBridge for CLiFF-LHMP~\cite{zhu2023cliff}, and through mode adapters for PPT~\cite{lin2024ppt}, GDTS~\cite{sun2025gdts}, Social Informer~\cite{jiang2026socialinformer}, and PECNet~\cite{mangalam2020pecnet}. This lets the prediction-side repair sit between different generators and the finite hypothesis sets exposed downstream.

On the complete Edinburgh online protocol with 3,719 test trajectories, DSR reduces MIF weighted ADE and FDE by \textbf{13.36\% and 13.30\%} with the same 64 maintained and returned hypotheses. Paired host integrations also reduce both metrics for CLiFF, PPT, causal GDTS, Social Informer, and PECNet. The gains therefore do not depend on a single recursive filter or trajectory generator.

This paper makes three contributions:
\begin{itemize}
  \item We identify support allocation under a fixed hypothesis budget as a distinct failure mode of multimodal prediction, covering both complete mode loss and under-allocation before extinction.
  \item We introduce DSR, a bounded causal repair rule that turns current candidate evidence into target mode counts, preserves active support, and moves surplus slots toward deficient modes. We establish support preservation, bounded intervention, and exact contraction toward the target allocation.
  \item We evaluate the same repair principle across a destination-particle filter, a flow-map bridge, and learned finite-set predictors, with cross-scene, component, support-event, sensitivity, and runtime analyses to separate where the gains come from and what they cost.
\end{itemize}
\begin{figure*}[!t]
    \centering
    \includegraphics[width=\textwidth]{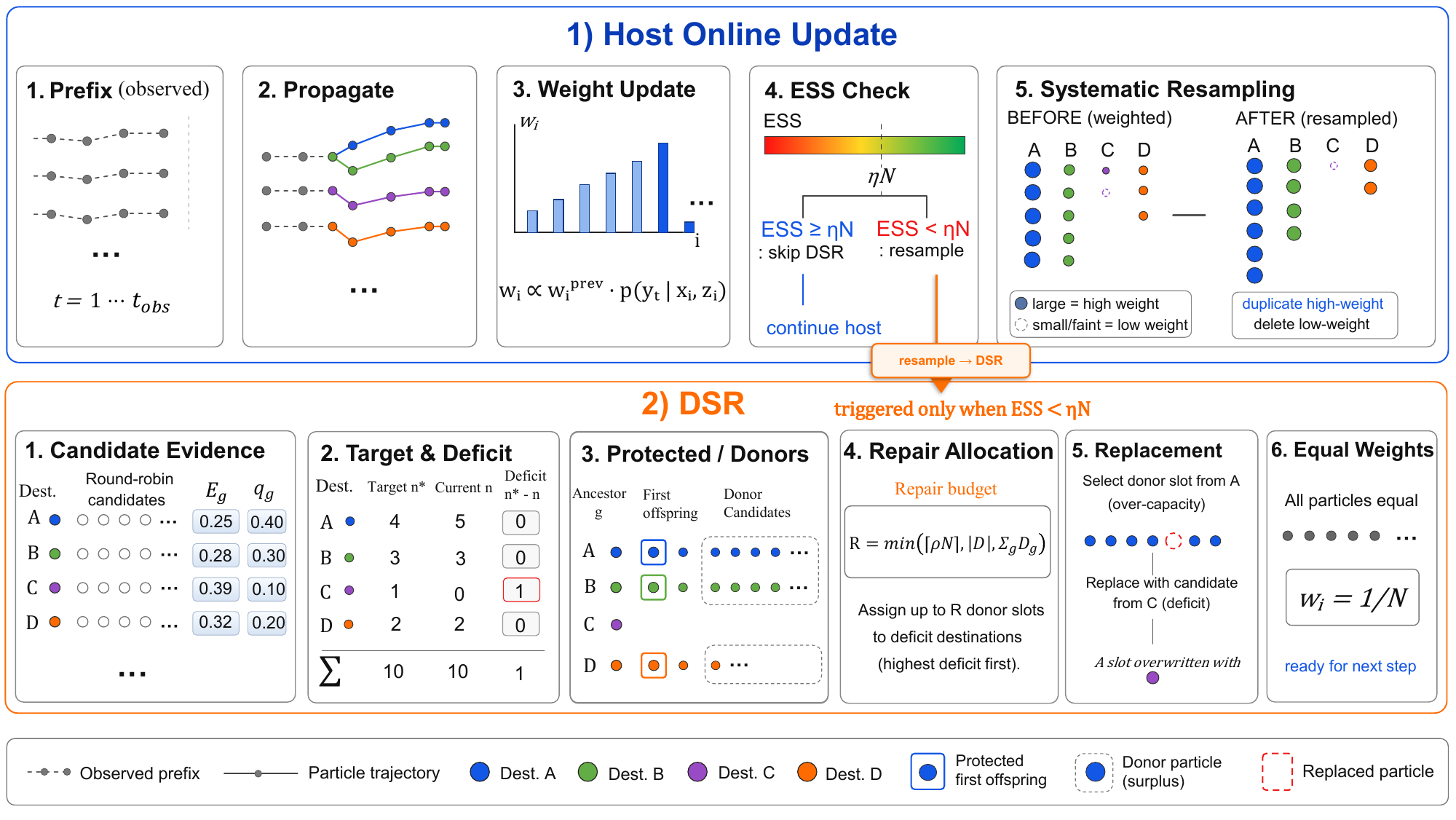}
    \caption{
    Host update and DSR pipeline. The host first propagates and weights its maintained hypotheses and then checks the effective sample size. When $\mathrm{ESS}<\eta N$, DSR maps evidence from a temporary destination-stratified candidate bank to target counts, protects active-mode representatives, replaces only unprotected surplus hypotheses, and returns $N$ equally weighted hypotheses to the host.
    }
    \label{fig:method}
    \vspace{-2mm}
\end{figure*}
\section{Related Work}
\subsection{Multimodal and Goal-Conditioned Trajectory Prediction}

Pedestrian forecasting has progressed from recurrent social-interaction models to increasingly expressive multimodal generators. Social LSTM models interactions through recurrent social pooling~\cite{alahi2016sociallstm}, while Social GAN introduces adversarial generation to produce multiple socially plausible futures~\cite{gupta2018socialgan}. Trajectron++ combines dynamic interaction graphs, latent variables, and agent dynamics~\cite{salzmann2020trajectron}, and SGCN learns sparse spatial--temporal interaction graphs~\cite{shi2021sgcn}. More recent diffusion models, including MID and LED, model stochastic futures through iterative denoising and provide expressive multimodal trajectory distributions~\cite{gu2022mid,mao2023led}.

A complementary line of work structures long-horizon uncertainty around endpoints or goals. PECNet first predicts a distant endpoint and conditions the intermediate trajectory on that endpoint~\cite{mangalam2020pecnet}, while Y-Net decomposes long-term forecasting into goals, waypoints, and paths~\cite{mangalam2021ynet}. PPT similarly incorporates destination prediction into a progressive learning framework, using short-term and destination pretext tasks before full trajectory prediction~\cite{lin2024ppt}. GDTS explicitly uses estimated goals to guide diffusion and introduces tree sampling for efficient multimodal generation~\cite{sun2025gdts}. Social Informer instead combines an Informer-based predictor with interaction modeling and an adaptive variance mechanism to represent stochastic pedestrian motion~\cite{jiang2026socialinformer}.

These methods improve how a model learns or samples a conditional trajectory distribution. DSR acts after the host exposes modes and conditional candidates. It reallocates the maintained and returned hypotheses across those modes. Goal-conditioned and multimodal predictors therefore serve as DSR hosts, while DSR leaves their generators unchanged.

\subsection{Recursive Intention and Long-Horizon Prediction}

Recursive probabilistic predictors maintain temporal continuity by updating latent motion intent as observations arrive. Goal-Directed Pedestrian Prediction represents destinations as latent variables and recursively estimates their probabilities~\cite{rehder2015goal}, while multi-hypothesis filtering maintains alternative pedestrian intentions during sequential inference~\cite{particke2018improvements}. MIF-WLSTM combines a mutable intention filter over predefined destination regions with a Warp LSTM trajectory decoder~\cite{huang2020mif}. Its mutation mechanism can introduce random replacements, but it does not construct an evidence-derived target count for each destination, restrict donation to target-surplus hypotheses, or explicitly protect every surviving resampling lineage.

CLiFF-LHMP samples long-horizon motion from learned flow-field dynamics~\cite{zhu2023cliff}. CLiFF exposes no particle ancestry, so our integration keeps its generated paths outside the DSR particle state. DSR instead maintains a finite goal-support state, and GoalBridge uses the repaired distribution to reweight the original CLiFF proposals. Lineage-aware filters provide ancestry-based protection; other predictors expose host-defined modes and protected representatives.

The focus of DSR is therefore narrower than constructing a complete recursive predictor. Propagation, observation processing, resampling or hypothesis generation, and trajectory decoding remain host operations. DSR intervenes only in the finite allocation between those stages.

\subsection{Particle Rejuvenation, Resampling, and Diverse Hypothesis Selection}

The finite-sample behavior motivating DSR connects to sample impoverishment in Sequential Monte Carlo~\cite{gordon1993bootstrap,doucet2001sequential}. Auxiliary particle filters use look-ahead information to improve proposal selection~\cite{pitt1999auxiliary}, resample--move methods apply Markov transitions after resampling~\cite{gilks2001moving}, and KLD sampling adapts particle count to approximation complexity~\cite{fox2001kld}. Mode-preserving and deterministic resampling retain distributional structure or reduce impoverishment~\cite{morelande2011mode,li2012deterministic}. Entropy-regularized optimal-transport resampling differentiably redistributes particle mass~\cite{corenflos2021dpf}.

DSR differs in both intervention point and objective. It leaves the host propagation, likelihood update, ESS criterion, and resampling rule unchanged. After the host has produced a finite set, DSR treats the number of slots assigned to each semantic mode as an explicit resource. It first protects the support that must survive, releases only target-surplus hypotheses outside that protected set, and reallocates a bounded number of those slots toward evidence-supported deficits. The resulting operation is not presented as an invariant Markov transition or an unbiased posterior correction; it is a controlled repair of the finite representation.

Output-diversification methods address another related problem. Likelihood-Based Diverse Sampling searches for separated high-likelihood trajectories from a pretrained distribution~\cite{ma2021lds}, while NPSN learns purposive samples for a frozen stochastic predictor~\cite{bae2022npsn}. Such methods improve coverage when constructing a new output set from a fixed observation window. DSR operates on an instantiated finite support state and reasons about categorical counts, protected representatives, and surplus-to-deficit reallocation. Diverse sampling controls where a generator places hypotheses; DSR controls their semantic allocation in the maintained set.

\section{Method}
Fig.~\ref{fig:method} summarizes the complete intervention between the host resampling step and the next online update.
\subsection{Finite Sets Expose Allocation Error}

At each prediction step, the host maintains $N$ hypotheses
\begin{equation}
\mathcal{S}=\{(x_i,z_i)\}_{i=1}^{N},
\end{equation}
where $x_i$ is a continuous state and $z_i\in\{1,\ldots,G\}$ is a destination or host-defined mode. Let
\begin{equation}
n_g=\sum_{i=1}^{N}\mathbf{1}[z_i=g]
\end{equation}
be the number of hypotheses in mode $g$.

DSR redistributes the $N$ maintained slots across plausible modes. It requires a mode-conditioned candidate sampler $Q(x\mid g)$, a causal compatibility energy $E(x,g;y_{1:t})$, and a protected set. For particle-filter hosts, DSR protects the first offspring of each surviving resampling ancestor. Hosts without ancestry protect one representative of every active mode. This second interface guarantees categorical support preservation; it makes no ancestry claim.

\subsection{Candidate Evidence Defines Target Counts}

At a repair step, DSR draws a temporary candidate bank stratified across modes. The bank exists only for the current allocation decision; DSR discards every candidate that does not enter the maintained set. For candidates $\mathcal{C}_g$ assigned to mode $g$, we compute
\begin{equation}
s_g=
\operatorname{LSE}_{c\in\mathcal{C}_g}
\left[-\beta E(c,g;y_{1:t})\right],
\qquad
\mathbf{q}=\operatorname{softmax}(\mathbf{s}).
\end{equation}
Here $q_g$ measures the current evidence for allocating finite support to mode $g$; it is not assumed to be a calibrated destination posterior.

We convert the real-valued allocation $N\mathbf{q}$ to integer target counts $n_g^\star$ with largest-remainder rounding, such that
\begin{equation}
\sum_{g=1}^{G}n_g^\star=N,
\qquad
\left|n_g^\star-Nq_g\right|<1.
\end{equation}
Only plausible underrepresented modes are eligible for repair:
\begin{equation}
d_g=(n_g^\star-n_g)_+\,\mathbf{1}[q_g\geq\tau],
\end{equation}
while a mode can donate at most
\begin{equation}
c_g=(n_g-n_g^\star)_+
\end{equation}
hypotheses.

\subsection{Surplus Donation Preserves Support}

DSR releases slots only from \textbf{unprotected surplus hypotheses}. It ranks eligible donors by decreasing energy and removes the least prefix-compatible hypotheses first. We bound the number of replacements by
\begin{equation}
R=
\min\left(
\left\lceil\rho N\right\rceil,
|D|,
\sum_{g=1}^{G}d_g
\right),
\end{equation}
where $D$ is the eligible donor set and $\rho$ is the maximum repair ratio.

The greedy allocator assigns the $R$ slots to deficient modes. With $r_g$ replacements already assigned to mode $g$, it assigns the next slot to
\begin{equation}
g^\star=
\underset{g:\,r_g<d_g}{\arg\max}\;
q_g\,d_g\,(d_g-r_g).
\end{equation}
This favors modes that are both plausible and below their target allocation.

DSR ranks candidates within each recipient mode by increasing energy and directly replaces the selected donors. The pair $(q_g,\tau)$ determines mode plausibility, while energy orders donor removal and candidate insertion. For particle-filter hosts, DSR operates on the finite set produced after the host likelihood update and resampling.

\subsection{Repair Contracts the Allocation Error}

Two finite-set properties follow from protected donor selection and surplus-to-deficit replacement.

\noindent\textbf{Proposition 1 (Support safety and bounded intervention).}
DSR modifies at most $\lceil\rho N\rceil$ hypotheses. If the protected set contains at least one representative of every active mode, then
\begin{equation}
\operatorname{supp}_z(\mathcal{S})
\subseteq
\operatorname{supp}_z(\mathcal{S}').
\end{equation}
For lineage-aware particle filters, protecting the first offspring of every distinct resampling ancestor additionally preserves at least one unchanged offspring of each surviving ancestor.

\noindent\textbf{Proposition 2 (Allocation contraction).}
Define
\begin{equation}
\Phi(\mathbf{n})=
\frac{1}{2}\left\|\mathbf{n}-\mathbf{n}^\star\right\|_1.
\end{equation}
Because every replacement moves one slot from a target-surplus mode to a target-deficient mode,
\begin{equation}
\Phi(\mathbf{n}')=\Phi(\mathbf{n})-R.
\end{equation}
Thus every DSR intervention moves the finite categorical allocation exactly $R$ steps toward its evidence-driven target.

MIF-WLSTM provides the lineage-aware instance through its destination particles and resampling ancestors. CLiFF uses a repaired goal-support state to guide its original proposals. PPT, GDTS, Social Informer, and PECNet use the support-safe interface: their adapters assign host hypotheses to modes and protect one representative of each active mode. Each host defines its candidate generator and compatibility energy; all hosts share the DSR allocation operator.
\begin{table*}[t]
\caption{Results on Edinburgh. Lower is better. We include historical AOE/FOE values marked with $\dagger$ as context because they follow a different protocol. NLL comparisons are paired within host; for CLiFF, we measure the DSR reduction relative to GoalBridge.}
\label{tab:main_results}
\centering
\resizebox{\textwidth}{!}{%
\begin{tabular}{lccccc}
\toprule
Predictor & Configuration & ADE/AOE $\downarrow$ & FDE/FOE $\downarrow$ & NLL $\downarrow$ & DSR Reduction \\
\midrule
Social Force~\cite{huang2020mif,helbing1995social}$^\dagger$ & - & 3.1240 & 3.9090 & - & - \\
LSTM~\cite{huang2020mif,hochreiter1997long}$^\dagger$ & - & 2.1320 & 3.0050 & - & - \\
Social LSTM~\cite{alahi2016sociallstm,huang2020mif}$^\dagger$ & - & 1.5240 & 2.5100 & - & - \\
Attention LSTM~\cite{huang2020mif,fernando2018soft}$^\dagger$ & - & 0.9860 & 1.3110 & - & - \\
Social GAN~\cite{gupta2018socialgan,huang2020mif}$^\dagger$ & 340 outputs & 1.0420 & 2.0880 & - & - \\
\midrule
MIF~\cite{huang2020mif} & $N=64$ & 0.6594 & 1.2420 & 5.1041 & - \\
MIF~\cite{huang2020mif} & $N=340$ & 0.6051 & 1.1404 & 3.9635 & - \\
\textbf{MIF + DSR} & $N=64$ & \textbf{0.5713} & \textbf{1.0768} & \textbf{3.7024} & \textbf{13.36\% / 13.30\%} \\
\midrule
CLiFF~\cite{zhu2023cliff} & $K=64$ & 0.5542 & 1.1194 & 3.2814 & - \\
CLiFF + GoalBridge & $K=64$ & 0.5319 & 1.0638 & 3.2021 & - \\
\textbf{CLiFF + GoalBridge + DSR} & $K=64$ & \textbf{0.5264} & \textbf{1.0493} & \textbf{3.1481} & \textbf{1.03\% / 1.36\%} \\
\midrule
PPT~\cite{lin2024ppt} & $H=20$ & 1.0931 & 2.4961 & 1.5351 & - \\
\textbf{PPT + DSR} & $H=20$ & \textbf{1.0024} & \textbf{2.2842} & \textbf{1.4359} & \textbf{8.30\% / 8.49\%} \\
Causal GDTS~\cite{sun2025gdts} & $H=20$ & 0.4925 & 0.9615 & 1.0163 & - \\
\textbf{Causal GDTS + DSR} & $H=20$ & \textbf{0.4563} & \textbf{0.8947} & \textbf{0.9853} & \textbf{7.36\% / 6.95\%} \\
Social Informer~\cite{jiang2026socialinformer} & $H=20$ & 0.4951 & 0.9683 & 1.0100 & - \\
\textbf{Social Informer + DSR} & $H=20$ & \textbf{0.4513} & \textbf{0.8874} & \textbf{0.9765} & \textbf{8.85\% / 8.36\%} \\
PECNet~\cite{mangalam2020pecnet} & $H=20$ & 0.4916 & 1.0554 & \textbf{1.8110} & - \\
\textbf{PECNet + DSR} & $H=20$ & \textbf{0.3709} & \textbf{0.7675} & 2.3484 & \textbf{24.55\% / 27.27\%} \\
\bottomrule
\end{tabular}%
}
\end{table*}
\section{Experiment}
\subsection{Experimental Setup}

\subsubsection{Dataset and Online Protocol}
We evaluate DSR on the Edinburgh Informatics Forum dataset~\cite{majecka2009statistical} with the online prediction protocol released with MIF-WLSTM~\cite{huang2020mif}. We represent trajectories at 10~Hz and use 34 square $1.5\,\mathrm{m}\times1.5\,\mathrm{m}$ boundary regions as destination categories. The test split contains 3,719 trajectories, and each run produces 86,404 online updates. Both prediction and correction windows contain 20 positions. Prediction starts after a 10-position warm-up and advances every two positions.

We report three-seed means for every method in the unified protocol. For each host, we compare the same model and prediction configuration with and without DSR. The evaluation emphasizes paired within-host changes because the host architectures use different probability parameterizations. All comparative conclusions reported below are supported by paired statistical analysis; complete confidence intervals and test outputs will be released together with the code.

MIF uses $N=64$ maintained destination particles in the main comparison. CLiFF returns $K=64$ proposals and connects to the repaired destination state through GoalBridge. PPT~\cite{lin2024ppt}, causal GDTS~\cite{sun2025gdts}, Social Informer~\cite{jiang2026socialinformer}, and PECNet~\cite{mangalam2020pecnet} use the support-safe interface. Their adapters associate host hypotheses with endpoint or destination modes and protect one representative of each active mode.

The main MIF configuration uses $C=68$, $\beta=10$, $\tau=0.01$, and $\rho=0.10$. Candidate evidence determines the target allocation, energy ranks surplus donors and recipient candidates, and each repair replaces at most seven of the 64 maintained hypotheses. Host-specific adapters retain the same allocation rule.

\subsubsection{Metrics}
We report weighted average displacement error (wADE) and weighted final displacement error (wFDE),
\begin{align}
\mathrm{wADE}
&=
\sum_{k=1}^{K}\omega_k\frac{1}{H}
\sum_{h=1}^{H}
\left\|\hat{\mathbf{y}}^{(k)}_h-\mathbf{y}_h\right\|_2,\\
\mathrm{wFDE}
&=
\sum_{k=1}^{K}\omega_k
\left\|\hat{\mathbf{y}}^{(k)}_H-\mathbf{y}_H\right\|_2,
\end{align}
where we normalize the output weights $\omega_k$ before evaluation. We report negative log-likelihood (NLL) as a secondary probabilistic score and compare it only within each host because the predictors use different probability parameterizations. NLL is evaluated under each host's own post-integration output distribution. Because DSR targets finite-set support allocation rather than the host's probabilistic scoring objective, lower displacement error need not imply a lower NLL.

\subsection{DSR Improves Paired Hosts}

Table~\ref{tab:main_results} compares DSR with its paired host baselines. We mark historical MIF-WLSTM results with $\dagger$ and include them only as context because they follow another protocol. We report relative DSR reductions only for matched host pairs.

\subsubsection{DSR Improves the Lineage-Aware Host}
On MIF, DSR reduces wADE from 0.6594 to 0.5713 and wFDE from 1.2420 to 1.0768 with the same $N=64$ maintained and returned particles. These changes equal relative reductions of 13.36\% and 13.30\%. The repaired $N=64$ model also outperforms the $N=340$ MIF reference, whose wADE and wFDE are 0.6051 and 1.1404. Complete DSR therefore outperforms a larger maintained set under the proposal and computation costs reported below.

\subsubsection{Complete DSR Transfers Across Hosts}
The improvement transfers beyond lineage-aware MIF. PPT reduces wADE and wFDE by 8.30\% and 8.49\%. Causal GDTS improves by 7.36\%/6.95\%, while Social Informer improves by 8.85\%/8.36\%. These models expose no resampling ancestry, so their adapters protect mode representatives. The paired gains show that complete DSR transfers across finite-set predictors with different architectures.

PECNet shows the largest displacement improvement, reducing wADE from 0.4916 to 0.3709 and wFDE from 1.0554 to 0.7675, corresponding to reductions of 24.55\% and 27.27\%. The complete NLL column decreases for MIF, CLiFF+GoalBridge, PPT, causal GDTS, and Social Informer; PECNet is the sole paired exception, increasing from 1.8110 to 2.3484. This divergence is consistent with DSR's objective: reallocating a finite support set can improve geometric trajectory coverage without necessarily improving the host-specific probabilistic score.

For CLiFF, the strict comparison is between GoalBridge and GoalBridge+DSR. DSR further reduces wADE/wFDE by 1.03\%/1.36\%. We do not attribute the larger difference between the original CLiFF output and GoalBridge+DSR entirely to DSR because GoalBridge itself modifies the proposal weights.

\subsection{Cross-Scene Generalization}

Table~\ref{tab:eth_ucy} evaluates MIF, CLiFF, and PPT on the five ETH/UCY scenes. Each entry reports a three-seed mean from a paired frozen-host comparison. MIF and CLiFF follow the destination-filter online protocol, while PPT uses its official frozen scene checkpoints with shared raw-track 8-observation/12-future windows. We compare each host only with its paired DSR integration because these protocols induce different absolute error scales.

\begin{table}[t]
\caption{Paired ETH/UCY results over three seeds. Each cell reports Host$\to$DSR; lower is better. UNIV denotes \texttt{stu03} for MIF and CLiFF.}
\label{tab:eth_ucy}
\centering
\scriptsize
\setlength{\tabcolsep}{2.0pt}
\resizebox{\columnwidth}{!}{%
\begin{tabular}{llcc}
\toprule
Host & Scene & wADE $\downarrow$ & wFDE $\downarrow$ \\
\midrule
MIF & ETH   & $1.2456\to\mathbf{1.2255}$ & $2.6604\to\mathbf{2.5983}$ \\
    & HOTEL & $0.4133\to\mathbf{0.4066}$ & $0.9036\to\mathbf{0.8825}$ \\
    & UNIV  & $0.7292\to\mathbf{0.7277}$ & $1.5781\to\mathbf{1.5699}$ \\
    & ZARA1 & $0.5258\to\mathbf{0.5185}$ & $1.1513\to\mathbf{1.1306}$ \\
    & ZARA2 & $0.4222\to\mathbf{0.4171}$ & $0.9280\to\mathbf{0.9134}$ \\
\midrule
CLiFF & ETH   & $1.1324\to\mathbf{1.1205}$ & $2.2120\to\mathbf{2.1076}$ \\
      & HOTEL & $0.4761\to\mathbf{0.4565}$ & $0.9777\to\mathbf{0.9683}$ \\
      & UNIV  & $0.7668\to\mathbf{0.7546}$ & $1.6127\to\mathbf{1.6062}$ \\
      & ZARA1 & $0.5585\to\mathbf{0.5536}$ & $1.1591\to\mathbf{1.1492}$ \\
      & ZARA2 & $0.4396\to\mathbf{0.4287}$ & $0.9193\to\mathbf{0.9130}$ \\
\midrule
PPT & ETH   & $4.2873\to\mathbf{4.2848}$ & $4.3242\to\mathbf{4.2958}$ \\
    & HOTEL & $1.8756\to\mathbf{1.8532}$ & $1.8856\to\mathbf{1.7762}$ \\
    & UNIV  & $1.6133\to\mathbf{1.6001}$ & $2.1627\to\mathbf{2.0534}$ \\
    & ZARA1 & $\mathbf{2.2856}\to2.2858$ & $1.8381\to\mathbf{1.7274}$ \\
    & ZARA2 & $1.5583\to\mathbf{1.5396}$ & $1.7013\to\mathbf{1.5604}$ \\
\bottomrule
\end{tabular}%
}
\end{table}

DSR lowers wFDE in all 15 host--scene pairs and lowers mean wADE in 14. The sole exception is PPT on ZARA1, where wADE changes from 2.2856 to 2.2858 while wFDE falls by 6.02\%. PPT yields a 5.16\% macro-average wFDE reduction across the five scenes. These results extend the Edinburgh finding across new scenes and two host interfaces.

\subsection{Ablation of DSR Components}

We ablate four design choices on MIF with $N=64$ while keeping the remaining evaluation configuration fixed. Setting $\rho=0$ disables repair entirely. The second variant removes lineage protection together with the surplus-only donor constraint. The final two variants replace the energy-based donor ranking or candidate ranking with random selection, respectively.

\begin{table}[t]
\caption{Component ablation on MIF with $N=64$. Lower is better.}
\label{tab:ablation}
\centering
\resizebox{\columnwidth}{!}{%
\begin{tabular}{lcc}
\toprule
Variant & wADE $\downarrow$ & wFDE $\downarrow$ \\
\midrule
\textbf{Full DSR} & \textbf{0.5713} & \textbf{1.0768} \\
Zero repair ($\rho=0$) & 0.69391 & 1.30667 \\
w/o lineage + surplus protection & 0.58088 & 1.09413 \\
Random donor & 0.59654 & 1.09956 \\
Random candidate & 0.58372 & 1.09875 \\
\bottomrule
\end{tabular}%
}
\end{table}

Disabling repair causes by far the largest deterioration, increasing wADE/wFDE by 21.46\%/21.35\% relative to Full DSR, which confirms that support repair is the main source of the gain. Removing lineage protection together with surplus-only donation also degrades performance to 0.58088/1.09413, indicating that the two protection mechanisms are effective jointly. Replacing energy-based donor ranking with random donor selection increases wADE/wFDE to 0.59654/1.09956, corresponding to degradations of 4.42\%/2.11\%. Random candidate selection yields 0.58372/1.09875, or 2.17\%/2.04\% worse than Full DSR. These results show that energy-based ranking contributes beyond mode-level repair, with donor ranking having the larger effect on average trajectory accuracy.

\subsection{Displacement Gains Persist After Support Repair}

Three support events test whether complete DSR behaves as the intended support-repair mechanism. Immediate revival records a mode that regains support at the repair step. Historical recovery records a mode that had zero support earlier and becomes active again. Pre-extinction increase records additional support assigned to a mode before it disappears. Random Reset is matched to DSR in intervention trigger and replacement count $R$, but selects donor slots and recipient candidates uniformly at random from the corresponding eligible pools rather than using target-deficit allocation or energy-based ranking.

\begin{table}[t]
\caption{Support-repair rates on MIF with $N=64$. Gain is the cluster-paired difference between method-specific rates in percentage points (pp).}
\label{tab:mechanism}
\centering
\resizebox{\columnwidth}{!}{%
\begin{tabular}{lccc}
\toprule
Mechanism & Random Reset & DSR & Gain (pp) \\
\midrule
Immediate revival & 2.94\% & 24.11\% & 21.17 \\
Historical recovery & 3.15\% & 22.21\% & 19.07 \\
Pre-extinction increase & 1.98\% & 22.72\% & 20.74 \\
\bottomrule
\end{tabular}%
}
\end{table}

\begin{figure*}[t]
    \centering
    \includegraphics[width=\textwidth]{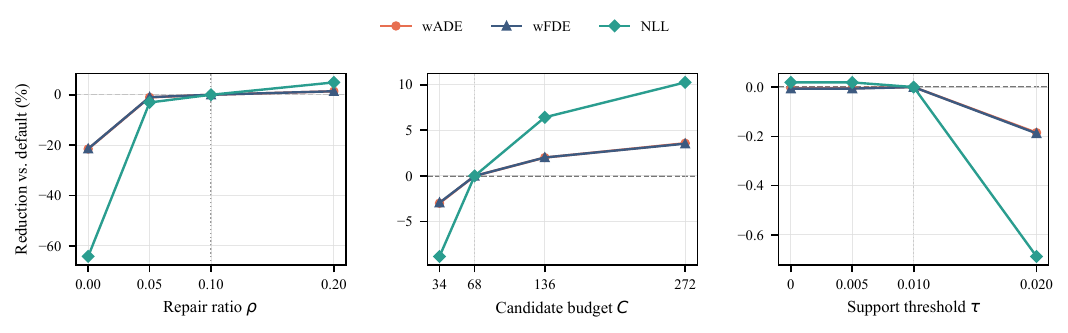}
    \caption{Sensitivity on MIF Edinburgh. Each panel changes one parameter while fixing the others at $\rho=0.10$, $C=68$, and $\tau=0.01$. Values show reduction relative to the default, so positive values indicate lower error. Dotted vertical lines mark the default settings.}
    \label{fig:sensitivity}
\end{figure*}
Table~\ref{tab:mechanism} uses 11,157 seed--trajectory clusters and 2,000 bootstrap repetitions. We use the true destination only to label posthoc support events; DSR receives no future labels. Each method induces its own eligible-event set, and the paired bootstrap matches seed--trajectory clusters instead of individual events. DSR raises every support-repair rate by at least 19.07~pp over Random Reset. Active-mode stratification localizes the displacement gains: DSR reduces wADE by 13.69\% with 1--2 active modes, 16.47\% with 3--4, and 4.96\% with at least five. Complete DSR therefore improves every bucket and yields its largest reduction when several alternatives compete for a small maintained set.

The intervention remains sparse. DSR invokes on 22.33\% of updates and performs a nonzero repair on 21.63\% of all updates. An active repair replaces a mean of 6.737 hypotheses and a median of seven, or 10.53\% and 10.94\% of the maintained set. Later predictions retain lower displacement error: after immediate revival, the mean DSR-minus-host wADE remains $-0.0784$ at $+1$ update, $-0.1056$ at $+5$, and $-0.1257$ at $+10$. wFDE follows the same pattern at $-0.1476$, $-0.2011$, and $-0.2428$. Paired cluster-bootstrap intervals support the same conclusions; the complete intervals and test outputs will be released with the code.

\subsection{Sensitivity to Repair Capacity and Support Threshold}

Fig.~\ref{fig:sensitivity} varies the repair ratio $\rho$, candidate budget $C$, and support threshold $\tau$ on MIF. The figure reports the relative reduction against the default setting; positive values favor the varied setting. Setting $\rho=0$ removes repair and increases wADE, wFDE, and NLL by 21.5\%, 21.4\%, and 64.2\% relative to the default. Increasing $\rho$ from 0.10 to 0.20 and increasing $C$ from 68 to 272 further reduce all three metrics, with higher runtime and more candidate scoring. The default values therefore balance intervention size and cost instead of maximizing offline accuracy. Results remain nearly unchanged for $\tau\in[0,0.01]$, while $\tau=0.02$ raises wADE and wFDE by about 0.2\%.

\subsection{Accuracy--Runtime Trade-off}

\begin{figure}[t]
    \centering   
    \includegraphics[width=1\columnwidth]{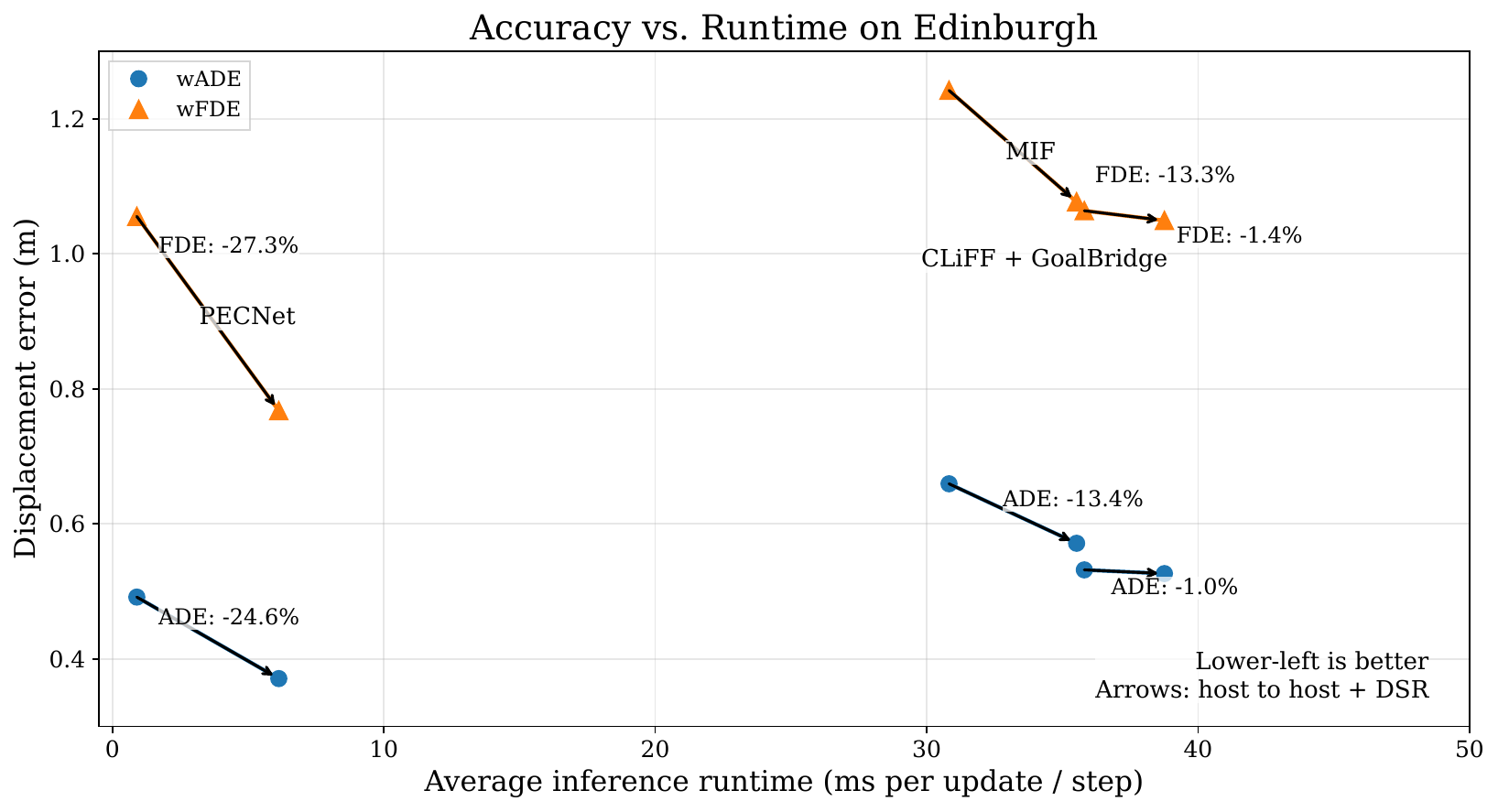}
    \caption{Accuracy--runtime trade-off on Edinburgh for MIF, CLiFF+GoalBridge, and PECNet. Each arrow connects a host configuration to its DSR-enhanced counterpart; lower-left is better. DSR reduces both wADE and wFDE for all three hosts while increasing average inference runtime by a model-dependent amount.}
    \label{fig:accuracy_runtime}
    \vspace{-2mm}
\end{figure}

Fig.~\ref{fig:accuracy_runtime} summarizes the accuracy--runtime trade-off of DSR on three representative Edinburgh hosts. For MIF, average inference runtime increases from 30.823~ms to 35.523~ms per update (15.25\%), while wADE and wFDE decrease by 13.36\% and 13.30\%. For CLiFF+GoalBridge, runtime increases from 35.807~ms to 38.758~ms (8.24\%), with corresponding wADE/wFDE reductions of 1.03\%/1.36\%. PECNet has a much lighter baseline: runtime rises from 0.90~ms to 6.13~ms per step, an absolute increase of 5.23~ms, while wADE/wFDE decrease by 24.55\%/27.27\%.

\section{Discussion and Conclusion}

The experiments support the central premise of DSR: under a fixed prediction budget, explicitly repairing support allocation improves the finite trajectory set without changing the host generator. The component ablation identifies repair as the dominant source of gain, the support-event analysis links the improvement to the intended restoration mechanism, and the cross-scene and runtime results show how the effect transfers and what additional computation it requires.More broadly, DSR shows that finite-set support allocation can serve as an explicit, lightweight control layer between prediction and downstream decision making.

DSR remains a prediction-side intervention rather than a complete planning solution, and its effectiveness depends on meaningful mode definitions and compatible candidate generation. Our evaluation is limited to pedestrian trajectory prediction with fixed repair hyperparameters; closed-loop robot experiments are left for future work. Within that scope, DSR provides a practical way to make the finite hypothesis set passed to downstream decision making better represent plausible human futures without increasing its maintained size.

\bibliographystyle{IEEEtran}
\bibliography{references}

@article{helbing1995social,
  author={Helbing, Dirk and Moln{\'a}r, P{\'e}ter},
  title={Social Force Model for Pedestrian Dynamics},
  journal={Physical Review E},
  volume={51},
  number={5},
  pages={4282--4286},
  year={1995},
  doi={10.1103/PhysRevE.51.4282}
}

@article{hochreiter1997long,
  author={Hochreiter, Sepp and Schmidhuber, J{\"u}rgen},
  title={Long Short-Term Memory},
  journal={Neural Computation},
  volume={9},
  number={8},
  pages={1735--1780},
  year={1997},
  doi={10.1162/neco.1997.9.8.1735}
}

@inproceedings{alahi2016sociallstm,
  author={Alahi, Alexandre and Goel, Kratarth and Ramanathan, Vignesh and Robicquet, Alexandre and Fei-Fei, Li and Savarese, Silvio},
  title={Social {LSTM}: Human Trajectory Prediction in Crowded Spaces},
  booktitle={Proceedings of the IEEE Conference on Computer Vision and Pattern Recognition},
  pages={961--971},
  year={2016},
  doi={10.1109/CVPR.2016.110}
}

@inproceedings{gupta2018socialgan,
  author={Gupta, Agrim and Johnson, Justin and Fei-Fei, Li and Savarese, Silvio and Alahi, Alexandre},
  title={Social {GAN}: Socially Acceptable Trajectories With Generative Adversarial Networks},
  booktitle={Proceedings of the IEEE Conference on Computer Vision and Pattern Recognition},
  pages={2255--2264},
  year={2018},
  doi={10.1109/CVPR.2018.00240}
}

@article{fernando2018soft,
  author={Fernando, Tharindu and Denman, Simon and Sridharan, Sridha and Fookes, Clinton},
  title={Soft+Hardwired Attention: An {LSTM} Framework for Human Trajectory Prediction and Abnormal Event Detection},
  journal={Neural Networks},
  volume={108},
  pages={466--478},
  year={2018},
  doi={10.1016/j.neunet.2018.09.002}
}

@inproceedings{salzmann2020trajectron,
  author={Salzmann, Tim and Ivanovic, Boris and Chakravarty, Punarjay and Pavone, Marco},
  title={Trajectron++: Dynamically-Feasible Trajectory Forecasting With Heterogeneous Data},
  booktitle={Proceedings of the European Conference on Computer Vision},
  pages={683--700},
  year={2020},
  doi={10.1007/978-3-030-58523-5_40}
}

@inproceedings{shi2021sgcn,
  author={Shi, Liushuai and Wang, Le and Long, Chengjiang and Zhou, Sanping and Zhou, Mo and Niu, Zhenxing and Hua, Gang},
  title={{SGCN}: Sparse Graph Convolution Network for Pedestrian Trajectory Prediction},
  booktitle={Proceedings of the IEEE/CVF Conference on Computer Vision and Pattern Recognition},
  pages={8994--9003},
  year={2021}
}

@inproceedings{mangalam2020pecnet,
  author={Mangalam, Karttikeya and Girase, Harshayu and Agarwal, Shreyas and Lee, Kuan-Hui and Adeli, Ehsan and Malik, Jitendra and Gaidon, Adrien},
  title={It Is Not the Journey but the Destination: Endpoint Conditioned Trajectory Prediction},
  booktitle={Proceedings of the European Conference on Computer Vision},
  pages={759--776},
  year={2020},
  doi={10.1007/978-3-030-58536-5_45}
}

@inproceedings{mangalam2021ynet,
  author={Mangalam, Karttikeya and An, Yang and Girase, Harshayu and Malik, Jitendra},
  title={From Goals, Waypoints \& Paths to Long Term Human Trajectory Forecasting},
  booktitle={Proceedings of the IEEE/CVF International Conference on Computer Vision},
  pages={15233--15242},
  year={2021}
}

@inproceedings{gu2022mid,
  author={Gu, Tianpei and Chen, Guangyi and Li, Junlong and Lin, Chunze and Rao, Yongming and Zhou, Jie and Lu, Jiwen},
  title={Stochastic Trajectory Prediction via Motion Indeterminacy Diffusion},
  booktitle={Proceedings of the IEEE/CVF Conference on Computer Vision and Pattern Recognition},
  pages={17113--17122},
  year={2022}
}

@inproceedings{mao2023led,
  author={Mao, Weibo and Xu, Chenxin and Zhu, Qi and Chen, Siheng and Wang, Yanfeng},
  title={Leapfrog Diffusion Model for Stochastic Trajectory Prediction},
  booktitle={Proceedings of the IEEE/CVF Conference on Computer Vision and Pattern Recognition},
  pages={5517--5526},
  year={2023}
}

@inproceedings{ma2021lds,
  author={Ma, Yecheng Jason and Inala, Jeevana Priya and Jayaraman, Dinesh and Bastani, Osbert},
  title={Likelihood-Based Diverse Sampling for Trajectory Forecasting},
  booktitle={Proceedings of the IEEE/CVF International Conference on Computer Vision},
  pages={13279--13288},
  year={2021}
}

@inproceedings{bae2022npsn,
  author={Bae, Inhwan and Park, Jin-Hwi and Jeon, Hae-Gon},
  title={Non-Probability Sampling Network for Stochastic Human Trajectory Prediction},
  booktitle={Proceedings of the IEEE/CVF Conference on Computer Vision and Pattern Recognition},
  pages={6477--6487},
  year={2022}
}

@inproceedings{rehder2015goal,
  author={Rehder, Eike and Kloeden, Horst},
  title={Goal-Directed Pedestrian Prediction},
  booktitle={Proceedings of the IEEE International Conference on Computer Vision Workshops},
  pages={139--147},
  year={2015},
  doi={10.1109/ICCVW.2015.28}
}

@inproceedings{particke2018improvements,
  author={Particke, Florian and Hiller, Markus and Feist, Christian and Thielecke, J{\"o}rn},
  title={Improvements in Pedestrian Movement Prediction by Considering Multiple Intentions in a Multi-Hypotheses Filter},
  booktitle={Proceedings of the IEEE/ION Position, Location and Navigation Symposium},
  pages={209--215},
  year={2018}
}

@mastersthesis{majecka2009statistical,
  author={Majecka, Barbara},
  title={Statistical Models of Pedestrian Behaviour in the Forum},
  school={School of Informatics, University of Edinburgh},
  address={Edinburgh, U.K.},
  year={2009}
}

@article{huang2020mif,
  author={Huang, Zhe and Hasan, Aamir and Shin, Kazuki and Li, Ruohua and Driggs-Campbell, Katherine},
  title={Long-Term Pedestrian Trajectory Prediction Using Mutable Intention Filter and Warp {LSTM}},
  journal={IEEE Robotics and Automation Letters},
  volume={6},
  number={2},
  pages={542--549},
  year={2021},
  doi={10.1109/LRA.2020.3047731}
}

@inproceedings{zhu2023cliff,
  author={Zhu, Yufei and Rudenko, Andrey and Kucner, Tomasz P. and Palmieri, Luigi and Arras, Kai O. and Lilienthal, Achim J. and Magnusson, Martin},
  title={{CLiFF-LHMP}: Using Spatial Dynamics Patterns for Long-Term Human Motion Prediction},
  booktitle={Proceedings of the IEEE/RSJ International Conference on Intelligent Robots and Systems},
  pages={3795--3802},
  year={2023},
  doi={10.1109/IROS55552.2023.10342031}
}

@article{gordon1993bootstrap,
  author={Gordon, Neil J. and Salmond, David J. and Smith, Adrian F. M.},
  title={Novel Approach to Nonlinear/Non-Gaussian Bayesian State Estimation},
  journal={IEE Proceedings F---Radar and Signal Processing},
  volume={140},
  number={2},
  pages={107--113},
  year={1993},
  doi={10.1049/ip-f-2.1993.0015}
}

@book{doucet2001sequential,
  editor={Doucet, Arnaud and de Freitas, Nando and Gordon, Neil},
  title={Sequential Monte Carlo Methods in Practice},
  publisher={Springer},
  address={New York, NY, USA},
  year={2001},
  doi={10.1007/978-1-4757-3437-9}
}

@article{pitt1999auxiliary,
  author={Pitt, Michael K. and Shephard, Neil},
  title={Filtering via Simulation: Auxiliary Particle Filters},
  journal={Journal of the American Statistical Association},
  volume={94},
  number={446},
  pages={590--599},
  year={1999},
  doi={10.1080/01621459.1999.10474153}
}

@article{gilks2001moving,
  author={Gilks, Walter R. and Berzuini, Carlo},
  title={Following a Moving Target---Monte Carlo Inference for Dynamic Bayesian Models},
  journal={Journal of the Royal Statistical Society: Series B (Statistical Methodology)},
  volume={63},
  number={1},
  pages={127--146},
  year={2001},
  doi={10.1111/1467-9868.00280}
}

@inproceedings{fox2001kld,
  author={Fox, Dieter},
  title={{KLD}-Sampling: Adaptive Particle Filters},
  booktitle={Advances in Neural Information Processing Systems},
  volume={14},
  pages={713--720},
  year={2001}
}

@inproceedings{morelande2011mode,
  author={Morelande, Mark R. and Zhang, Alan M.},
  title={A Mode Preserving Particle Filter},
  booktitle={Proceedings of the IEEE International Conference on Acoustics, Speech and Signal Processing},
  pages={3984--3987},
  year={2011}
}

@article{li2012deterministic,
  author={Li, Tiancheng and Boli{\'c}, Miodrag and Djuri{\'c}, Petar M.},
  title={Deterministic Resampling: Unbiased Sampling to Avoid Sample Impoverishment in Particle Filters},
  journal={Signal Processing},
  volume={92},
  number={7},
  pages={1637--1645},
  year={2012},
  doi={10.1016/j.sigpro.2011.12.019}
}

@inproceedings{corenflos2021dpf,
  author={Corenflos, Adrien and Thornton, James and Deligiannidis, George and Doucet, Arnaud},
  title={Differentiable Particle Filtering via Entropy-Regularized Optimal Transport},
  booktitle={Proceedings of the 38th International Conference on Machine Learning},
  series={Proceedings of Machine Learning Research},
  volume={139},
  pages={2100--2111},
  year={2021}
}

@inproceedings{lin2024ppt,
  author={Lin, X. and Liang, T. and Lai, J. and Hu, J.-F.},
  title={Progressive Pretext Task Learning for Human Trajectory Prediction},
  booktitle={Proceedings of the European Conference on Computer Vision},
  year={2024}
}

@inproceedings{sun2025gdts,
  author={Sun, G. and Wang, S. and Zhu, L. and Liu, M. and Ma, J.},
  title={{GDTS}: Goal-Guided Diffusion Model with Tree Sampling for Multi-Modal Pedestrian Trajectory Prediction},
  booktitle={Proceedings of the IEEE/RSJ International Conference on Intelligent Robots and Systems},
  year={2025}
}

@article{jiang2026socialinformer,
  author={Jiang, Z. and Yang, R. and Ma, Y. and Qin, C. and Chen, X. and Wang, Z.},
  title={Social Informer: Pedestrian Trajectory Prediction by Informer With Adaptive Trajectory Probability Region Optimization},
  journal={IEEE Transactions on Cybernetics},
  volume={56},
  number={1},
  pages={15--28},
  year={2026}
}

\end{document}